\documentclass[10pt,twocolumn,letterpaper]{article}

\usepackage{iccv}              % To produce the CAMERA-READY version
\usepackage{amsmath}
\definecolor{iccvblue}{rgb}{0.21,0.49,0.74}
\usepackage[pagebackref,breaklinks,colorlinks,allcolors=iccvblue]{hyperref}

\title{U-StyDiT: Ultra-high Quality Artistic Style Transfer Using Diffusion Transformers}

\author{Zhanjie Zhang$^{1,2,*}$, Ao Ma$^{2,*}$, Ke Cao$^{2,*}$, Jing Wang$^{2}$, Shanyuan Liu$^{2}$, Yuhang Ma$^{2}$, \\Bo Cheng$^{2}$, Dawei Leng$^{2\dag}$, Yuhui Yin$^{2}$\\
{\tt\small $^*$Equal Contribution, $^\dag$Corresponding Authors}\\
$^1$College of Computer Science and Technology, Zhejiang University \\ $^2$360 AI Research \\
{cszzj}@zju.edu.cn, lengdawei@360.cn
}

\begin{document}

	\twocolumn[{%
	\renewcommand\twocolumn[1][]{#1}%
	\maketitle
	\begin{center}
		\centering
		\includegraphics[width=0.99\linewidth]{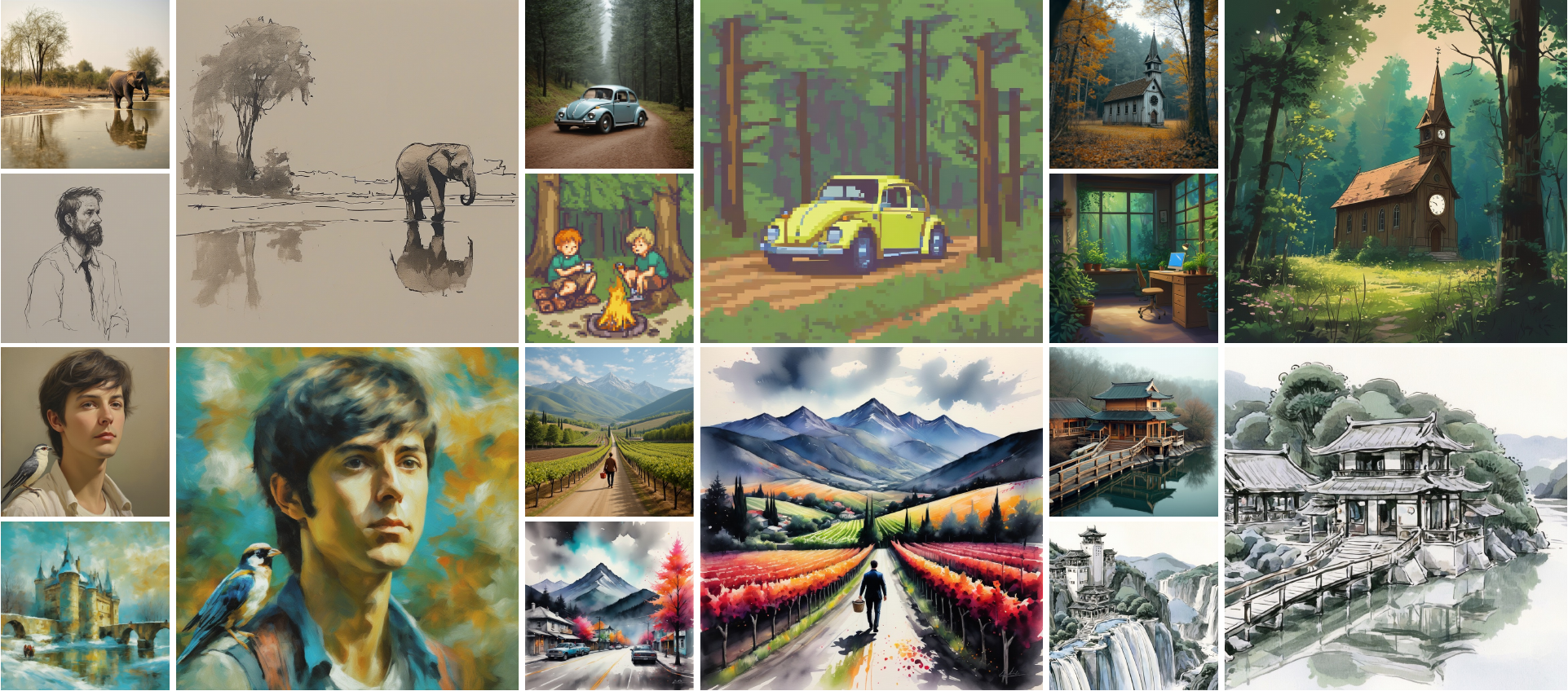}
		\captionof{figure}{Artistic style transfer results by the proposed U-StyDiT. Given a content and style image, our proposed U-StyDiT successfully produces ultra-high quality style transfer results that preserve the structure of the content image and the style information of the style image.}
		\label{image1}
	\end{center} 
}]

\begin{abstract}
Ultra-high quality artistic style transfer refers to repainting an ultra-high quality content image using the style information learned from the style image. Existing artistic style transfer methods can be categorized into style reconstruction-based and content-style disentanglement-based style transfer approaches. Although these methods can generate some artistic stylized images, they still exhibit obvious artifacts and disharmonious patterns, which hinder their ability to produce ultra-high quality artistic stylized images. To address these issues, we propose a novel artistic image style transfer method, U-StyDiT, which is built on transformer-based diffusion (DiT) and learns content-style disentanglement, generating ultra-high quality artistic stylized images. Specifically, we first design a Multi-view Style Modulator (MSM) to learn style information from a style image from local and global perspectives, conditioning U-StyDiT to generate stylized images with the learned style information. Then, we introduce a StyDiT Block to learn content and style conditions simultaneously from a style image.
Additionally, we propose an ultra-high quality artistic image dataset, Aes4M, comprising 10 categories, each containing 400,000 style images. This dataset effectively solves the problem that the existing style transfer methods cannot produce high-quality artistic stylized images due to the size of the dataset and the quality of the images in the dataset. Finally, the extensive qualitative and quantitative experiments validate that our  U-StyDiT can create higher quality stylized images compared to state-of-the-art artistic style transfer methods. To our knowledge, our proposed method is the first to address the generation of ultra-high quality stylized images using transformer-based diffusion.
\end{abstract}
\section{Introduction}
\label{sec:intro}
Given a content image, ultra-high quality artistic style transfer refers to transferring the learned style information onto this content image, generating an ultra-high quality artistic stylized image. Existing artistic
style transfer methods can be primarily categorized into two types: style reconstruction-based ~\cite{zhang2024artbank,flux-ipa,ye2023ip,wang2024instantstyle,hertz2024style,zhang2023inversion,zhang2025dyartbank,wang2024qihoo,cao2025relactrl} and content-style disentanglement-based style transfer approaches~\cite{zhang2024towards,wang2023stylediffusion,zhang2023prospect,xing2024csgo,gao2024styleshot}.

\begin{figure}[htb]
		\centering
		\includegraphics[width=1\columnwidth]{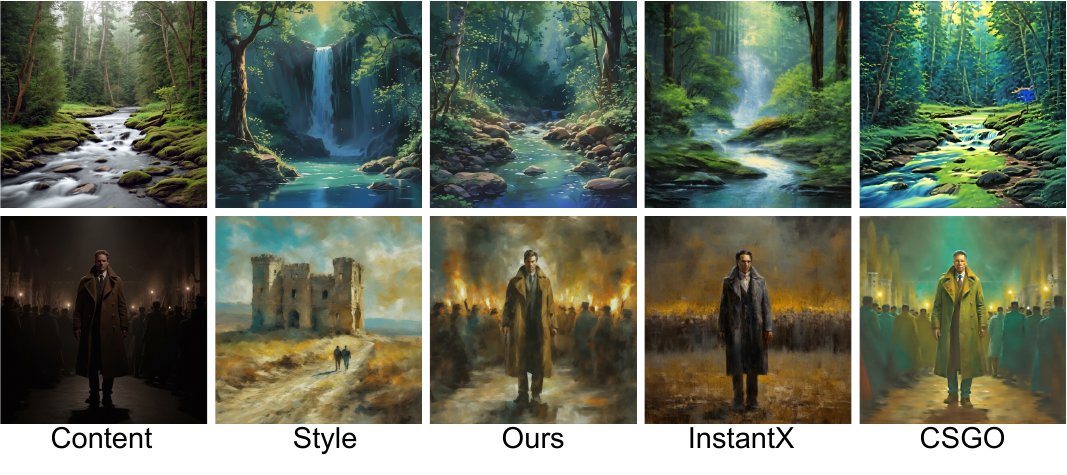} % Reduce the figure size so that it is slightly narrower than the column.
		\caption{We present some stylized image examples generated by our proposed U-StyDiT, InstantX~\cite{flux-ipa} and CSGO~\cite{xing2024csgo}. Existing image stylization methods fail to create ultra-high quality stylized images, introducing obvious artifacts and disharmonious patterns.  
		}
		\label{image2}
\end{figure}

More specifically, style reconstruction-based style transfer approaches usually train a style adapter only on style images to condition stable diffusion (SD)~\cite{rombach2022high,feng2024fancyvideo,liu2023bridge,ma2024hico} to generate desired images with the learned style information. For example, Ip-adapter~\cite{ye2023ip} introduced a style adapter on Unet-based stable diffusion to learn style information from a style image. Recently, FLUX~\cite{BlackForestLabs2024}, trained using DiT-based stable diffusion~\cite{peebles2023scalable}, has shown outstanding capability in producing high-quality images. InstantX trained a style adapter on FLUX to inject style information. However, during the implementation of style transfer, these methods use a Canny image~\cite{canny1986computational} as an additional condition to preserve the content structure of the stylized images. We argue that since the style condition and Canny are typically trained on separate style and content image datasets, this inevitably leads to information confusion when used simultaneously, introducing obvious artifacts and disharmonious patterns (e.g., in the $4^{th}$ col of Fig.~\ref{image2}).

The content-style disentanglement-based style transfer approaches simultaneously focus on training content and style conditions. 
For example, CSGO~\cite{xing2024csgo} created a triad dataset that includes content images, style images, and stylized images. Training the model to decouple the stylized images into the corresponding content and style conditions can facilitate the generation of stylized images using both content and style images in inference. StyleShot~\cite{gao2024styleshot} trained a Mixture-of-Expert (MoE)~\cite{riquelme2021scaling} module on a pre-trained content condition to extract multi-level style embeddings. However, although these methods learn the disentanglement of content and style to some degree, they still cannot produce ultra-high quality stylized images. We believe that the main reasons for this limitation are twofold: 1) Compared to transformer-based diffusion, the Unet-based diffusion has certain limitations in capturing long-range dependencies among patches, which hinders its ability to generate ultra-high quality stylized images. 2) Lack of  ultra-high quality artistic style images and effective methods for model to learn content-style disentanglement.

\begin{figure}[htb]
		\centering
		\includegraphics[width=1\columnwidth]{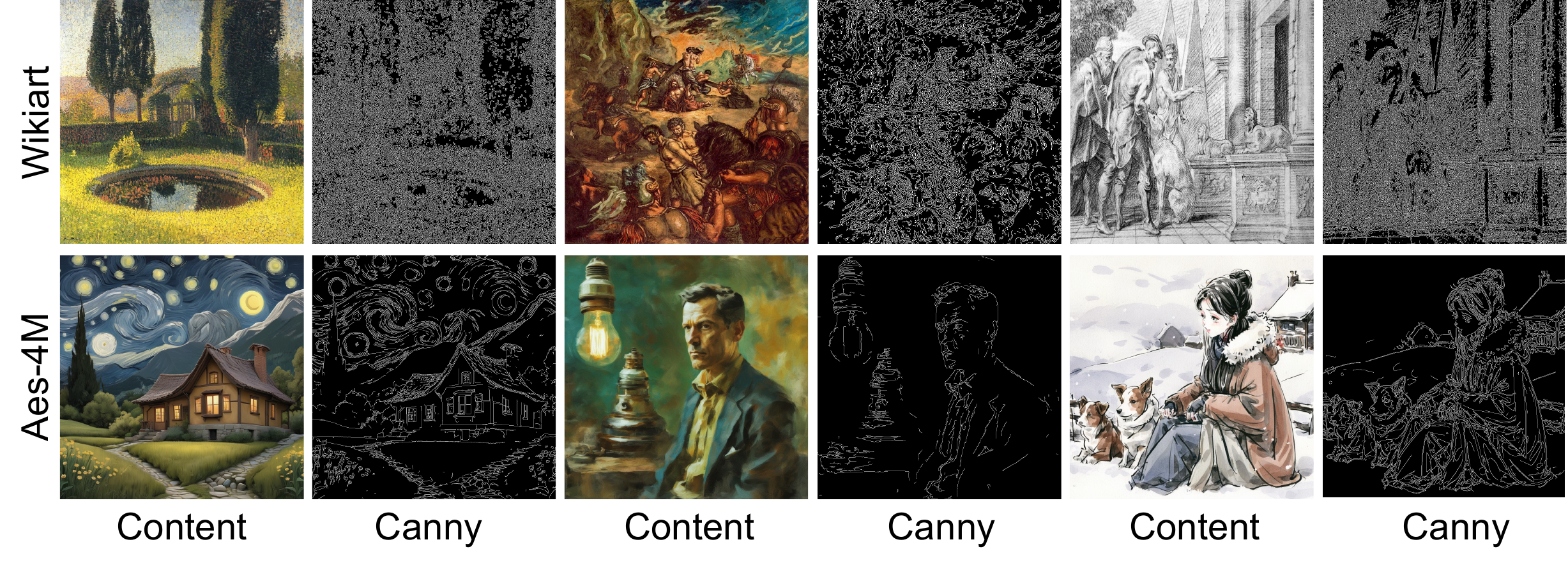} % Reduce the figure size so that it is slightly narrower than the column.
		\caption{Compared to Wikiart~\cite{wikiart}, Aes4M has clearer Canny images.
		}
		\label{image10}
        \vspace{-0.5em}
\end{figure}
To solve the above problems, we introduce a novel framework, U-StyDiT, to generate ultra-high quality stylized images. First, we focus on designing a modulator to inject style information into a transformer-based diffusion model~\cite{Shakker2024}. In detail, we propose a novel Multi-view Style Modulator (MSM) to extract style information from a style image from both local and global perspectives. Specifically, we simultaneously extract style information from the entire style image and multiple patches within the style image. Unlike previous methods~\cite{ye2024stylemaster,zhang2025lgast}, we do not aggressively discard style image patches that have lower similarity to the overall image; instead, we merge these patches at the token level and sample style information to reduce computational complexity. Furthermore, we introduce a StyDiT Block to replace the existing DiT block, addressing the challenge of learning content and style conditions simultaneously based on transformer-based stable diffusion. To our knowledge, this is the first paper that simultaneously trains content and style conditions on transformer-based stable diffusion to accomplish the task of generating ultra-high quality stylized images (e.g., We present some ultra-high quality stylized image samples generated by our method in Fig.~\ref{image1}). In addition, we find that existing methods typically use Canny~\cite{canny1986computational} to extract content conditions. However, Canny often encounters difficulties when applied to existing artistic images, such as those from Wikiart~\cite{wikiart}. This is because artistic images in Wikiart tend to have a high frequency of detailed texture, making it challenging to extract Canny images and style information from human-created artistic collections to train content and style conditions.
% This results in a tendency to use natural images to train canny conditions, while using artistic images to train style modulators. Since canny conditions and style modulators are trained separately on different datasets, it is often impossible to obtain satisfactory artistic stylized pictures in inferences. 
To this end, we propose an ultra-high quality artistic image dataset, Aes4M, which consists of 10 categories, with 400,000 artistic images in each category, totaling 4 million artistic images. Each artistic style image has high aesthetics and text-image consistency, as well as clear Canny image.
% This advantage is very helpful for us to train the style demodulator and canny conditions during the training process, and to synthesize ultra-high quality stylized images in inference.
In conclusion, we show our contributions as follows:
\begin{itemize}
\item We propose a novel artistic image style transfer
method, U-StyDiT, which is built on transformer-based
diffusion (DiT) and can generate ultra-high quality artistic stylized images. 
\item We propose a novel Multi-view Style Modulator (MSM) to extract style information from a style image from both local and global perspectives and introduce a StyDiT Block to simultaneously address the challenge of learning content and style conditions on transformer-based stable diffusion.
\item We build an ultra-high quality artistic image dataset, Aes4M, consisting of 10 categories, with 400,000 artistic images in each category, totaling 4 million. Unlike previous datasets, Aes4M possesses high aesthetics, high text-image consistency, and clear Canny images.
\item  Extensive qualitative and quantitative experiments verify the effectiveness of our proposed U-StyDiT compared to the state-of-the-art artistic style transfer methods.
\end{itemize}
%-------------------------------------------------------------------------
\section{Related Works}
\textbf{Style Reconstruction-based Style Transfer.}
Style reconstruction-based style transfer approaches refer to training a style adapter only on style images to condition stable diffusion to generate the desired images. For example, PerText~\cite{galimage} employed pseudo-words in the embedding space of a pre-trained text-to-image model as a specific concept to acquire style information from a style image. Inst~\cite{zhang2023inversion} learned a text embedding from a style image and used the learned text embedding to condition stable diffusion to generate the desired stylized image. Artbank~\cite{zhang2024artbank} introduced an implicit style condition to learn style information from multiple style images. StyleID~\cite{Chung_2024_CVPR} injected style information into stable diffusion by replacing the key and value of content features with the key and value of style features. Ip-adapter~\cite{ye2023ip} employed a style adapter to infuse style information into the Unet-based stable diffusion model, enabling it to learn global style information from a given style image. InstantStyle~\cite{wang2024instantstyle} found that in the Unet-based stable diffusion model, style information is only controlled by the specific layers. This phenomenon motivates InstantStyle to inject style information solely into the specific layers.
StyleAligned~\cite{hertz2024style} implemented minimal attention-sharing operation along the diffusion process and generated style-consistent stylized images. 

While these methods can effectively implement style transfer, they ignore learning the effective content conditions and fail to generate high quality stylized images. As a result, the generated stylized images often exhibit obvious artifacts and disharmonious patterns.

\textbf{Content-style Disentanglement-based Style Transfer.}
Content-style disentanglement-based style transfer approaches simultaneously learn content and style conditions from style images. For example, CSAST~\cite{kotovenko2019content} first introduced a triplet style loss to learn style variations within different style images and content disentanglement loss to lean the content structure to ensure that the stylization is not conditioned on the real input photo.
LSAST~\cite{zhang2024towards} proposed a step-aware and layer-aware prompt to learn style patterns from style images, effectively decoupling the content structure and style patterns of a style image. Stylediffusion~\cite{wang2023stylediffusion} introduced a CLIP-based~\cite{radford2021learning} style disentanglement loss to achieve content-style disentanglement from a style image. Prospect~\cite{zhang2023prospect} separated the content and style information from the step dimension within DDPM~\cite{ho2020denoising}. Styleshot~\cite{gao2024styleshot} employed a Mixture-of-Expert (MoE) module, which is trained on a pre-existing content condition to capture multi-level style embeddings. CSGO~\cite{xing2024csgo} created a triad dataset consisting of content, style, and stylized images. Training the model to separate the stylized image into its corresponding content and style images facilitates the generation of new stylized images during inference by utilizing both content and style images.

Although these methods take into account the disentanglement of content structure and style information, they are limited by the availability of high-quality datasets and the use of Unet-based diffusion models, which prevents them from generating ultra-high quality stylized images.

% \begin{table*}[htb]
% 		\footnotesize
% 		\centering
% 		\begin{tabular}{c|ccccccccccc}
% 			\hline  Number of Patches  & 1  & 2 & 3 & 4 & 5 & 6 & 7& 8& 9& 10 \\
% 			\hline  Parameters (M) $\downarrow$ & 18.5 &37.0 & 55.5 &74.0  & 92.5 & 111.0 & 129.50&148.00&166.50&185.0 \\
% 			GFLOPs  $\downarrow$ & 1.5  &6.2&13.9 & 24.9 & 39.8&57.7 & 78.6&102.6&129.6 &159.5 \\
% 			\hline  Parameters (M) $\downarrow$ & 18.5 &19.5 & 20.5 &21.5.0  & 22.5 & 23.5 & 24.5&25.5&26.5&27.5 \\
% 			GFLOPs  $\downarrow$ & 1.5  &6.2&13.9 & 24.9 & 39.8&57.7 & 78.6&102.6&129.6 &159.5 \\
% 			\hline
% 		\end{tabular}
% 		\caption{Quantitative comparison. The best result is signed in \textbf{bold}.}
% 		\label{comparisontable}
% 	\end{table*}
\begin{figure}[htb]
		\centering
		\includegraphics[width=1\columnwidth]{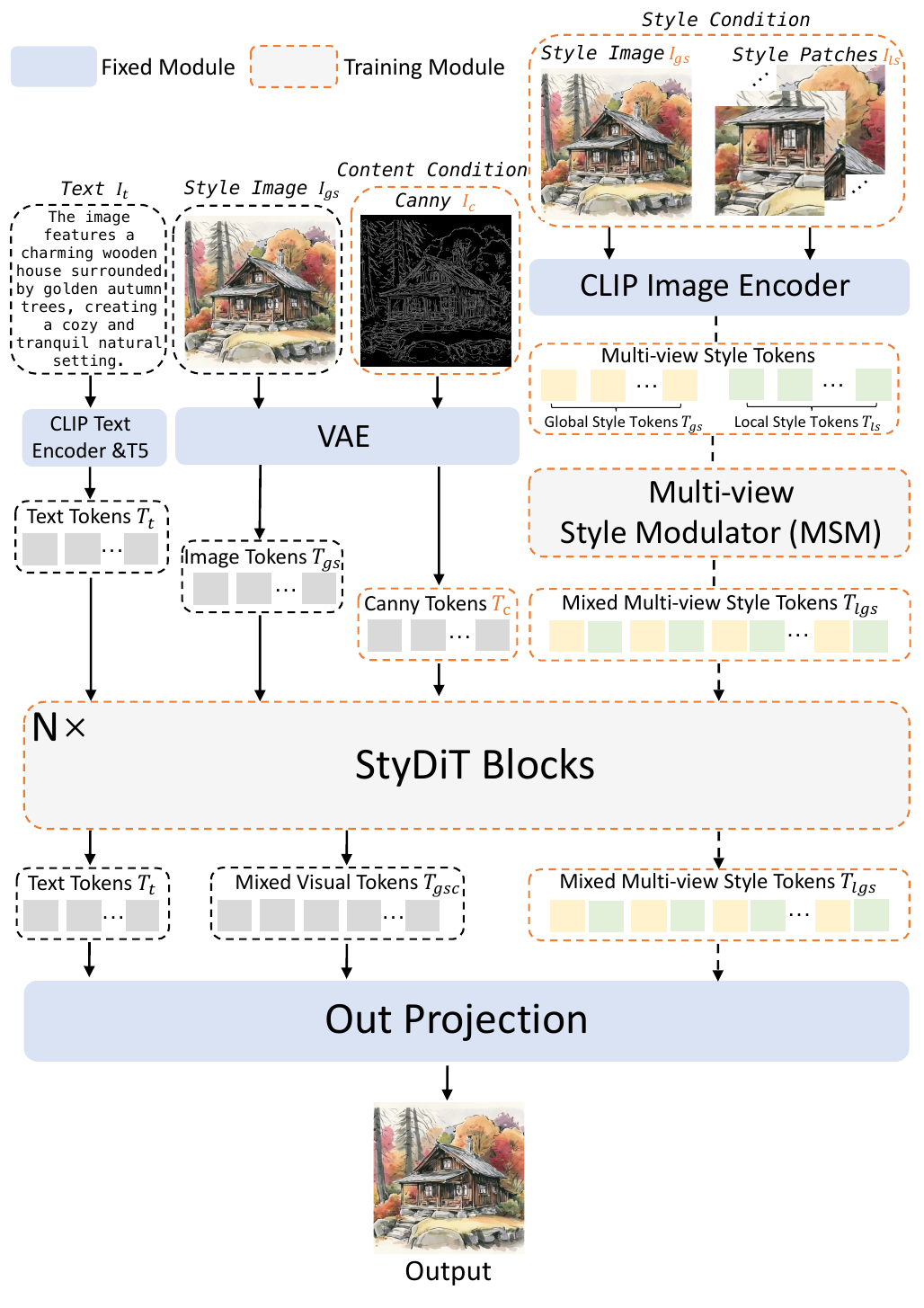} % Reduce the figure size so that it is slightly narrower than the column.
		\caption{The training pipeline of our U-StyDiT. Given an artistic image from Aes4M with a resolution of $1024\times 1024$, we randomly crop multiple  $512\times 512$ style patches $I_{ls}$. Then, we resize this artistic image to  $512\times 512$, obtaining style image $I_{gs}$. We use Chatglm~\cite{glm2024chatglm} to extract text $I_{t}$ from $I_{gs}$. When extracting Canny maps from the $I_{gs}$, the high threshold is set to 200, and the low threshold is set to 100.
		}
		\label{image3}
\end{figure}
\section{Methods}
Given a content image and a style image, U-StyDiT aims to transfer the style information from the style image onto the content image and generate ultra-high quality stylized images. 
% Specifically, U-StyDiT learns to decouple a style image into content and style conditions. 
Since our network is designed based on transformer-based diffusion and use FLUX.1-dev~\cite{BlackForestLabs2024} as the backbone, we call it U-StyDiT. 

Fig.~\ref{image3} shows the overview of training our U-StyDiT, which consists of two key ingredients: 1) a Multi-view Style Modulator (MSM) and 2) a StyDiT Block module. During the training phase, U-StyDiT learns to reconstruct a style image $I_{gs}$ based on the given text $I_{t}$, content condition, and style condition. Once U-StyDiT is fully trained, we can extract the style condition from the style images, and content condition from the content images to guide U-StyDiT to generate desired stylized images from the noise images (see the supplementary material for the inference pipeline).

\subsection{Multi-view Style Modulator}
\label{Multi-view Style Modulator}
\begin{figure}[htb]
		\centering
		\includegraphics[width=1\columnwidth]{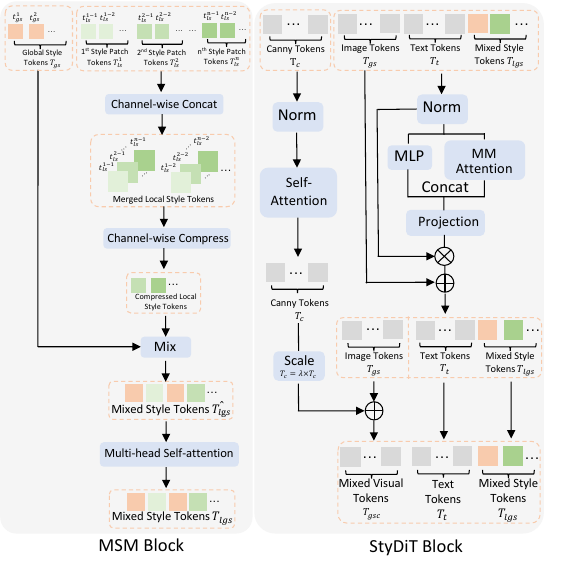} % Reduce the figure size so that it is slightly narrower than the column.
		\caption{The details of Multi-view Style Modulator and StyDiT.  
		}
		\label{image4}
\end{figure}

We believe that training the network structure with high resolution ($1024\times1024$) based on the existing large-scale data generally allows the model to produce higher quality images.
However, we find that full-tuning FLUX at high resolutions incurs additional computational burdens, such as increased computational complexity and memory requirements. To tackle this problem, we resize images with a resolution of $1024\times1024$ to $512\times512$ and then perform full-tuning of FLUX.  Directly resizing images from $1024\times1024$ to $512\times512$ can result in the loss of some style information. Therefore, we propose a Multi-view Style Modulator (MSM) that balances image resolution with computational demands.  Specifically, given a style image $I_{s}$ with a resolution of $1024\times1024$, we resize $I_{s}$ to obtain resized $I_{gs}$ with a resolution of $512\times512$, while also randomly cropping multiple style image patches \{$I^{1}_{ls}, I^{2}_{ls},...,I^{n}_{ls}$\} from $I_{s}$, each style image patch with a resolution of $512\times512$. 
Previous methods~\cite{ye2024stylemaster,zhang2025lgast} directly discard style image patches that have lower similarity to the overall image. They argue that the style information provided by these style image patches is not representative of the overall style image. We consider this naive approach to be highly inappropriate, as style patches with low similarity often capture the unique characteristics of the style image. Therefore, we use a Multi-view Style Modulator to learn style information from style images and style image patches.
Specifically, given a resized style image $I_{gs}$ and multiple cropped style image patches \{$I^{1}_{ls}, I^{2}_{ls},I^{3}_{ls},...$\}, we use transformer encoder to extract resized style image token $T_{gs}$ and style image patch tokens \{$T^{1}_{ls}, T^{2}_{ls},T^{3}_{ls},...$\}. Each style image patch is projected into a sequence of style patch tokens. (i.e., $T^{1}_{ls}=\{t^{1-1}_{ls},t^{1-2}_{ls},t^{1-3}_{ls},...\},T^{n}_{ls}=\{t^{n-1}_{ls},t^{n-2}_{ls},t^{n-3}_{ls},...\}$, n=10). In fact, we extracted a total of 10 style patches from the style image $I_{s}$. If we directly learned the style relationship between resized style image tokens and cropped style tokens, it would require additional computational complexity and memory. This is because we need to learn the relationships among $\{T_{gs},T^{1}_{ls},T^{2}_{ls},T^{3}_{ls},...,T^{10}_{ls}\}$. To this end, as shown in Fig.~\ref{image4}, we concatenate multiple cropped style image patch tokens from channel dimension, obtaining Merged Local Style Tokens $T_{ls}$ (i.e., $T_{ls}=\{[t^{1-1}_{ls},t^{2-1}_{ls},...,t^{n-1}_{ls}],[t^{2-1}_{ls},t^{2-1}_{ls},...,t^{n-2}_{ls}],... \}$,n=10). We redefine the process of obtaining Merged Local Style Tokens from multiple style image patch tokens as follows:
\begin{equation}
T_{ls} = Concat[T^{1}_{ls}, T^{2}_{ls},...,T^{n}_{ls}],
\end{equation}
where $Concat$ means concatenate different tokens from channel dimension. Then, we further compress $T_{ls}$, obtaining Compressed Local Style Tokens $\hat{T_{ls}}$. For example, $T^{1}_{ls}$ is a sequence of tokens, such as $T^{1}_{ls}= \{t^{1-1}_{ls},t^{2-1}_{ls},...,t^{n-1}_{ls}\}$. 
For $T^{1}_{ls}$, we use MLP to predict a sequence of weight $\alpha^{1}_{ls}=\{\alpha^{1-1}_{ls},\alpha^{2-1}_{ls},...,\alpha^{n-1}_{ls}\}$ (i.e., $\alpha^{1}_{ls}=MLP(T^{1}_{ls})$) to further compress as below:
\begin{equation}
\hat{T^{1}_{ls}} = \alpha^{1-1}_{ls}\times t^{1-1}_{ls} + \alpha^{2-1}_{ls}\times t^{2-1}_{ls}+...+\alpha^{n-1}_{ls}\times t^{n-1}_{ls}
\end{equation}
Then, we compress $T^{2}_{ls}$ to $T^{n}_{ls}$ using the same way, obtaining the Compressed Local Style Tokens: $\hat{T_{ls}}$ as below:
\begin{equation}
\hat{T_{ls}} = Compress(T_{ls}),
\end{equation}
where $\hat{T_{ls}}=\{t^{1-1}_{ls},t^{1-2}_{ls},t^{1-3}_{ls},...\}$.
Then, we can obtain Mixed Style Tokens $T_{lgs}$ via merging Global Style Tokens and Local Style Tokens from spatial dimension as below:
\begin{equation}
\hat{T_{lgs}} = Mix(\hat{T_{ls}},T_{gs}),
\end{equation}
Further, we use the Mixed Style Tokens $T_{lgs}$ to generate the Query $Q$, Key $K$ and Value $V$:
\begin{equation}
Q=\hat{T_{lgs}}W_q, \quad K=\hat{T_{lgs}} W_k, \quad V=\hat{T_{lgs}} W_v ,
\end{equation}
Then, the relationship between the mixed style tokens can be calculated via attention as below:
\begin{equation}
\begin{aligned}
\hat{T_{lgs}^{ }} & =\mathcal{F}_{\mathrm{MSA}}(Q, K, V)+Q, \\
\hat{T_{lgs}^{ }}  & =\mathcal{F}_{\mathrm{MSA}}\left(\hat{T_{lgs}^{ }} , K, V\right)+\hat{T_{lgs}^{ }},  \\
T_{lgs}^{ }& =\mathcal{F}_{\mathrm{FFN}}\left(\hat{T_{lgs}^{ }}\right)+\hat{T_{lgs}^{ }},
\end{aligned}
\end{equation}
where $\mathcal{F}_{\mathrm{MSA}}$ denotes multi-head self-attention~\cite{vaswani2017attention} and $\mathcal{F}_{\mathrm{FFN}}$ means forward fully connected network. 
We redefine the above process as below:
\begin{equation}
\label{MSM}
T_{lgs} = MSM(T_{gs},T^{1}_{ls},...,T^{n}_{ls}),n=10.
\end{equation}

\subsection{U-StyDiT Blocks}
\label{U-StyDiT Blocks}
FLUX.1-dev is a text-to-image model based on Diffusion Transformer (DiT)~\cite{peebles2023scalable,chen2023pixarta,tan2024ominicontrol} that is capable of generating ultra-high quality images.

The FLUX.1-dev architecture comprises a text encoder, a VAE~\cite{kingma2013auto}, and a DiT module, with the input consisting of two types of tokens: image tokens and text condition tokens. In FLUX.1-dev, each DiT block comprises layer normalization followed by Multi-Modal Attention (MMA)~\cite{esser2024scaling}. For image tokens $T_{i}$ and text $T_{t}$ tokens, the Multi-Modal Attention (MMA) projects them into queries (Q), keys (K), and values (V), facilitating the computation of attention:
\begin{equation}
\operatorname{MMA}\left(\left[T_{\mathrm{i}} ; T_{t} \right]\right)=\operatorname{Softmax}\left(\frac{Q K^{\top}}{\sqrt{d}}\right) V,
\end{equation}
where $[T_{\mathrm{i}} ; T_{t} ]$ means the concatenation of image and text tokens. Softmax denotes the Softmax function~\cite{bishop2006pattern}. Building upon the FLUX.1-dev~\cite{BlackForestLabs2024}, OminiControl~\cite{tan2024ominicontrol} introduces a conditional image token $T_{s}$ into the multi-modal attention computation to support the injection of conditional image information, as shown below:
\begin{equation}
\label{MMA}
\operatorname{MMA}\left(\left[T_{\mathrm{i}} ; T_{t} ; T_{\mathrm{s}}\right]\right)=\operatorname{Softmax}\left(\frac{Q K^{\top}}{\sqrt{d}}\right) V,
\end{equation}
where $[T_{\mathrm{i}} ; T_{t} ;T_{c} ]$ means the concatenation of image, text, conditional image tokens. Based on Eq.~\ref{MMA}, we replace conditional image information via mixed style tokens $T_{\mathrm{lgs}}$ to introduce style information as below:
\begin{equation}
[T_{gs};T_{t};T_{lgs}]=\operatorname{MMA}\left(\left[T_{\mathrm{gs}} ; T_{t} ; T_{\mathrm{lgs}}\right]\right)
\end{equation}
where $T_{gs}$ denotes images tokens from resized style image.
However, simply adding style information in this manner often leads to the injection of only style condition, making it difficult to learn both content and style conditions simultaneously.
To this end, we introduce a novel StyDiT Block, as shown in Fig.~\ref{image4}, which add Canny tokens $T_{c}$ to the image tokens $T_{gs}$, obtaining mixed visual tokens $T_{gsc}$ as below:
\begin{equation}
T_{gsc}= \lambda \times T_{c} + T_{gs}, \lambda \in [0,1]
\end{equation}
% two conditional images to tackle these issues. Specifically, we added two additional conditions based on FLUX.1-dev: content condition and style condition.  Given a resized style image and a canny image, we first use VAE to extract image tokens $T_{gs}$ and Canny tokens $T_{c}$, and use Eq.~\ref{MSM} to extract mixed style tokens $T_{lgs}$. Then, we use Multi-Modal Attention to learn the relationships among them:
% \begin{equation}
% \label{eq9}
% \operatorname{MMA}\left(\left[T_{\mathrm{gs}} ; T_{t} ; \lambda\times T_{\mathrm{c}};T_{\mathrm{lgs}}\right]\right)=\operatorname{Softmax}\left(\frac{Q K^{\top}}{\sqrt{d}}\right) V,
% \end{equation}
where $\lambda$ ($\lambda \in [0,1]$) denotes the control strength of the canny condition. 
We redefine above process as below:
\begin{equation}
\label{StyDiT}
[T_{gsc};T_{t};T_{lgs}]=\operatorname{StyDiT}\left(\left[T_{\mathrm{gs}} ; T_{t} ; T_{\mathrm{lgs}};\lambda \times T_{\mathrm{c}};\right]\right)
\end{equation}
Notably, StyDiT addresses how to learn both content and style conditions on transformer-based diffusion.
% Further, we redefine the above process as follows:
% \begin{equation}
% \label{eq2}
% \operatorname{U-Sty}=\operatorname{MMA}\left(\left[T_{\mathrm{gsc}} ; T_{t} ; \lambda\times T_{\mathrm{c}};T_{\mathrm{lgs}}\right]\right)
% \end{equation}
\begin{figure*}[htb]
		\centering
		\includegraphics[width=2.07\columnwidth]{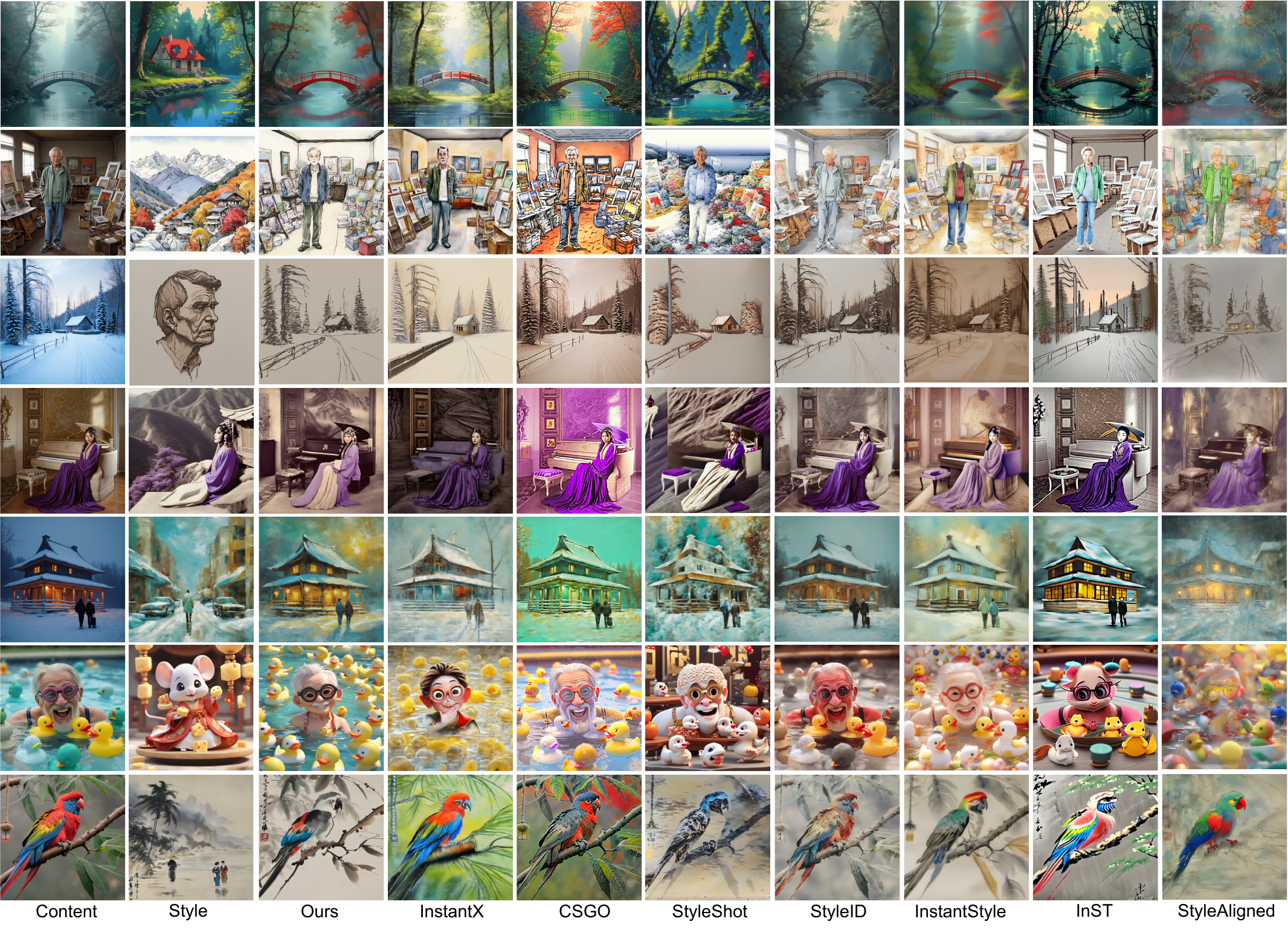} % Reduce the figure size so that it is slightly narrower than the column.
		\caption{We compare our method with the sate-of-the-art artistic style transfer method, including style reconstruction-based style transfer approaches (e.g., InstantX~\cite{flux-ipa}, StyleID~\cite{Chung_2024_CVPR}, InstantStyle~\cite{wang2024instantstyle}, InST~\cite{zhang2023inversion} and StyleAligned~\cite{hertz2024style}) and content-style disentanglement-based style transfer approaches (e.g., CSGO~\cite{xing2024csgo} and StyleShot~\cite{gao2024styleshot}).
		}
		\label{image5}
\end{figure*}
\subsection{Aes4M datasets}
Aes4M consists of 10 categories, with 400,000 artistic images in each category, totaling 4 million. Each image has a resolution of $1024\times1024$. To collect these images, we  collect 10 LoRA weights~\cite{hulora} from Civitai~\cite{civitai} that can generate high-quality style images that are highly relevant to practical business applications. We use these LoRA weights to control SD3.0, SD3.5, SDXL~\cite{rombach2022high}, and FLUX~\cite{BlackForestLabs2024} to produce high-quality stylized images. These primarily include 10 types of style information, such as oil, cartoon, Gufeng, pixel, paint, 3D, sketch, Peking opera, cute, and wash painting (Please refer to the supplementary materials for details). The step of building Aes4M is illustrated below:
\begin{itemize}

\item \textbf{Prompt Generation.} 
We first use GPT-4o to generate 1.7 million diverse descriptions. The text descriptions involve landscapes, objects etc. These text descriptions aim to depict the open world as comprehensively as possible.

\item \textbf{Style Image Synthesis.}
For each LoRA weight, we feed it into the corresponding diffusion model and generate 1.7 million images using 1.7 million text descriptions. We filter the generated images based on text-image consistency, aesthetic evaluation, and Canny-image similarity. After filtration, each category comprising approximately 400,000 style images.

\item \textbf{Text-image Consistency.} To ensure that the images in the Aes4M dataset have a high degree of text-image consistency, we used a text-image consistency evaluation model~\cite{clip-vit-large-patch14} to filter out images with a consistency score below 30.  

\item  \textbf{Aesthetic Evaluation.}
To ensure that each image in Aes4M has a high aesthetic score, we use an aesthetic assessment model~\cite{Schuhmann,xing2024diffsketcher} to filter the generated images, removing those with an aesthetic score below 7.

\item \textbf{Canny-image Similarity Constraint.}
Compared to the Wikiart dataset~\cite{wikiart}, our proposed Aes4M has clearer Canny information. We use CLIP Score~\cite{radford2021learning} to calculate the similarity between each image and its corresponding Canny image, and the similarity between each image and its canny image reached more than 0.67.

\end{itemize}

\begin{table*}[htb]
            \small
		\centering
		\begin{tabular}{c|c|cccccccc}
			\hline  Metric  & U-StyDiT & InstantX & CSGO & StyleShot &StyleID &InstantStyle & InST & StyleAligned \\
			\hline  SSIM $\uparrow$ & $\mathbf{0.421}$ &0.231 & 0.256 &0.153  & 0.376 & 0.145 & 0.312 &0.196\\
			Clip Score $\uparrow$ & $\mathbf{0.623}$ &0.551&0.560 & 0.615 & 0.552&0.609 & 0.493&0.540 \\
                Aesthetic Score $\uparrow$ & $\mathbf{6.974}$ &6.332&6.159 & 6.720 & 6.611&6.882 &5.805&6.161 \\
			  Deception Rate$\uparrow$ & $\mathbf{0.781}$  & 0.645 & 0.597 & 0.586 & 0.531 &0.563&0.493 & 0.540 \\
			Preference Score $\uparrow$& \textbf{59.3}* & 53.6/46.4 & 57.5/42.5 &58.3/41.7 & 64.7/35.3 & 60.1/39.9 &  61.4/38.6&59.9/40.1 \\
			\hline Time (Sec./Image) $\downarrow$ & 23.125 & 27.212s
        & 18.916 & 4.934s & 14.641s & 11.791s & 6.132s&27.623s \\
			\hline
		\end{tabular}
		\caption{Quantitative comparison. The best result is signed in \textbf{bold}.}
		\label{comparisontable}
\end{table*}

\section{Experiments}
\subsection{Implementation Details}
We use FLUX.1-dev~\cite{BlackForestLabs2024} as the base model and full-tune the base model. During the training process, we randomly selected style images from Aes4M and uniformly resized them to a resolution of $512\times512$ pixels. The U-StyDiT is trained with a batch size of
2 and gradient accumulation over 4 steps. The experiments are conducted on 32 NVIDIA A100 GPUs (80GB each). The model was trained for a total of 2 million iterations. Additionally, when extracting Canny maps from the style images, the high threshold is set to 200, and the low threshold is set to 100.

\subsection{Comparisons with SOTA Methods}
We compare our U-StyDiT against seven state-of-the-art methods~\cite{flux-ipa,Chung_2024_CVPR,wang2024instantstyle,zhang2023inversion,hertz2024style,xing2024csgo,gao2024styleshot} from both qualitative and quantitative comparisons.

\textbf{Qualitative Comparisons}
Comparison with SOTA artistic style transfer methods. We compare our method with the state-of-the-art artistic style transfer method, including style reconstruction-based style transfer approaches (e.g., InstantX~\cite{flux-ipa}, StyleID~\cite{Chung_2024_CVPR}, InstantStyle~\cite{wang2024instantstyle}, InST~\cite{zhang2023inversion} and StyleAligned~\cite{hertz2024style}) and content-style disentanglement-based style transfer approaches (e.g., CSGO~\cite{xing2024csgo} and StyleShot~\cite{gao2024styleshot}. As the representative of style reconstruction-based style transfer approaches, InstantX tends to generate some stylized images with blurred texture (e.g., $1^{st}$ and $5^{th}$ rows). StyleID fails to preserve the content structure (e.g., $2^{nd}$ row). InstantStyle sometimes produces stylized images with excessive style degree (e.g., $3^{rd}$ row). InST always fails to preserve the content structure of the input content image (e.g., $4^{th}$ and $6^{th}$ row). StyleAligned has some limitations in balancing content structure and style patterns and introduces disharmonious patterns (e.g., $5^{th}$ and $6^{th}$ row). As the representative of content-style disentanglement-based style transfer approaches, CSGO sometimes introduces obvious artifacts (e.g., $1^{st}$ and $2^{nd}$ rows). StyleShot generates stylized images with obvious artifacts and disharmonious patterns (e.g., $5^{th}$ row).  

Compared to the above methods, our proposed U-StyleDiT can generate ultra-high quality artistic stylized images without introducing obvious artifacts and disharmonious patterns.
\subsection{Quantitative Comparisons}
The average Structural Similarity Index (SSIM)~\cite{an2021artflow} to assess the content similarity between content image and stylized images. To compare our proposed method's superiority in content preservation against existing style transfer methods, we collected 50 content images and 20 style images, synthesizing 1,000 stylized images to calculate the average SSIM between the stylized images and their corresponding content images. As shown in the $2^{nd}$ row of Tab.~\ref{comparisontable}, the results demonstrate that our method exhibits a higher structural similarity between the stylized images and content images than other state-of-the-art style transfer methods.

CLIP score~\cite{radford2021learning} is usually used to evaluate the style similarity between the stylized and corresponding images. As shown in the $3^{rd}$ row of Tab.~\ref{comparisontable}, we calculated the average CLIP score between 1,000 stylized images and their corresponding style images. The results demonstrate that the images generated by our method exhibit a higher style consistency with the reference style images.

To further evaluate the quality of the images generated by our proposed method, we employed an aesthetic assessment model~\cite{Schuhmann,xing2024diffsketcher} to evaluate 1,000 stylized images comprehensively and calculated the average aesthetic score. As shown in $4^{th}$ row of Tab.~\ref{comparisontable}, our proposed method achieved the highest, 6.974, aesthetic score, which further demonstrates the superiority of our approach in generating high-quality images from the perspective of the assessment model.

\begin{figure*}[htb]
		\centering
		\includegraphics[width=2.07\columnwidth]{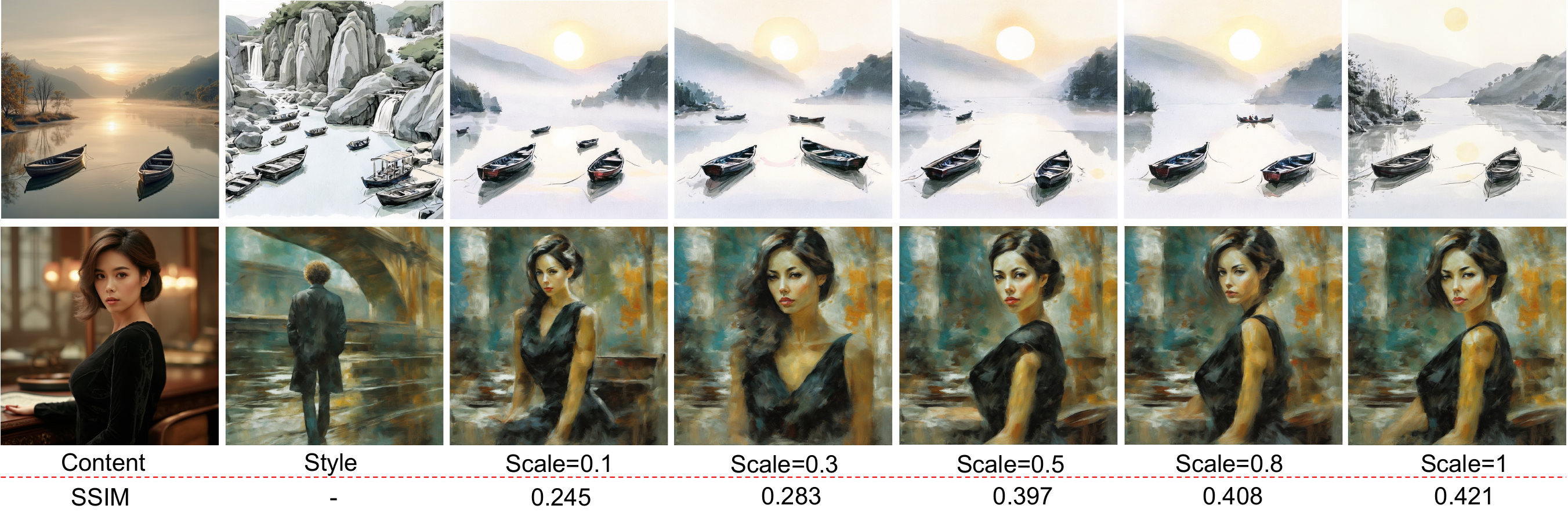} % Reduce the figure size so that it is slightly narrower than the column.
		\caption{The ablation study on the StyDiT. We scale the coefficient $\lambda$ (see Eq.~\ref{StyDiT}) from 0.1 to 1 to demonstrate content controllability.  Please zoom-in for better comparison. 
		}
		\label{image12}
\end{figure*}
\begin{figure}[htb]
		\centering
		\includegraphics[width=1\columnwidth]{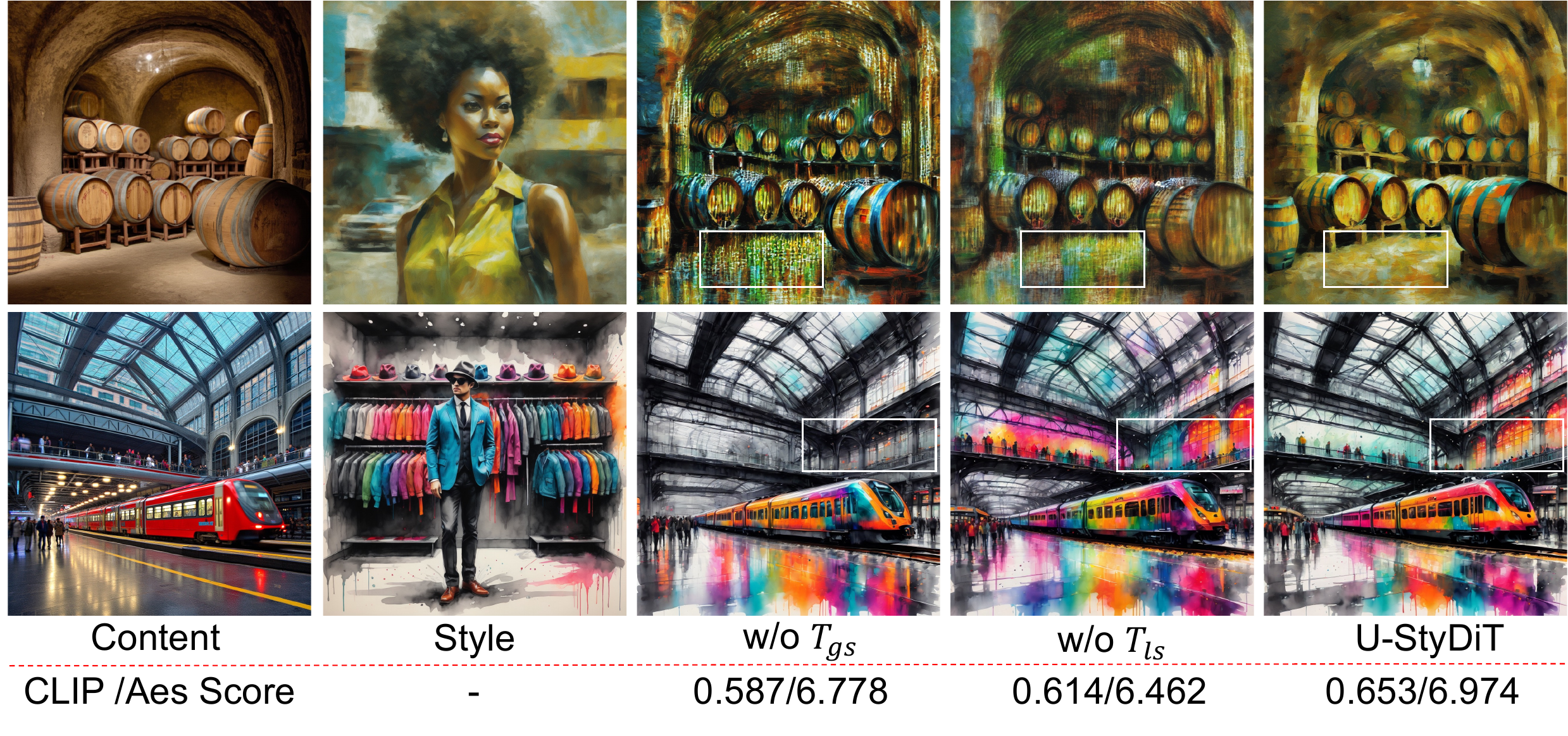} % Reduce the figure size so that it is slightly narrower than the column.
		\caption{The ablation study on the Multi-view Style Modulator. 
		}
		\label{image11}
\end{figure}
\textbf{User Study}
Although we have evaluated the generated stylized images using SSIM, CLIP Score, and Aesthetic Score, the assessment of stylized images is also highly influenced by subjective perspectives. Therefore, we conduct a Preference Score~\cite{chen2021dualast} to evaluate our method. Preference Score is commonly used to compare the popularity of two methods. Specifically, we randomly selected 100 pairs of content and style images. For each content and style image pair, we presented users with a stylized image generated by our method and a stylized image generated by other state-of-the-art methods. Users were asked to choose their preferred image based on the style pattern and content structure. We obtained 2,000 votes for each question from 40 users and displayed the percentage of votes supporting our method over the others as shown in the $4^{th}$ row of Tab.~\ref{comparisontable}. The higher score indicates that our method is preferred over the other methods. We can easily conclude that, compared to other style transfer methods, the stylized images generated by our method are more favored by users.

Additionally, we used deception scores~\cite{sanakoyeu2018style} to evaluate whether the synthesized images were likely to be perceived as human-created. We randomly chose 100 synthesized images for each method and asked 50 participants to guess whether the generated images were human-created. As shown in the $5^{th}$ row of Tab.~\ref{comparisontable}, 78\% of participants believe our generated images were human-created, suggesting that our proposed method can generate stylized images closer to human-created images.

\textbf{Timing Information}
As shown in the $7^{th}$ row of Tab.~\ref{comparisontable}, we present the inference time~\cite{wang2022aesust,zhang2023caster,zhang2025lgast} comparisons with a resolution of $512\times512$ pixels under A40. Due to the use of the more powerful FLUX.1-dev~\cite{BlackForestLabs2024}, which is based on a transformer diffusion structure, our method has certain drawbacks in inference time. However, it is undeniable that our approach can generate ultra-high-quality stylized images.

\subsection{Ablation Studies.}
\textbf{Multi-view Style Modulator.} We propose a Multi-view Style Modulator (MSM) to learn style information from the high-resolution images and balance computational demands. To verify the effectiveness of MSM, we remove global style tokens $T_{gs}$ and local style tokens $T_{ls}$ respectively to validate the effectiveness of our method. As shown in Fig.~\ref{image11}, w/o $T_{ls}$, local style information becomes confused ($1^{st}$ row, $4^{th}$ col), causing the same object to fail to preserve the same colors ($2^{nd}$ row, $4^{th}$ col). Additionally, the style similarity between the stylized images and the style images decreases (indicated by a lower CLIP score), and the aesthetic quality of the stylized images also decreases (reflected by a lower Aesthetic Score). W/o $T_{gs}$, the stylized images exhibit some obvious artifacts ($1^{st}$ row, $3^{th}$ col) and fail to achieve the desired stylization in some certain details ($2^{nd}$ row, $3^{th}$ col).

\textbf{StyDiT.}
The proposed StyDiT block can learn content and style conditions simultaneously. Once the StyDiT has been trained, we can control the structural similarity between the generated stylized image and the content image by controlling the coefficient $\lambda$ of the content condition.
As shown in Fig.~\ref{image12}, we scale the coefficient $\lambda$ from 0.1 to 1 to demonstrate content controllability. With a higher coefficient $\lambda$, the generated stylized images show higher structural similarity with the content image.

\section{Conclusion}
In this paper, we present a novel artistic image style transfer framework called U-StyleDiT, which is built on transformer-based diffusion and is capable of generating ultra-high quality artistic stylized images. Specifically, we design a Multi-view Style Modulator to learn style information from a style image from both local and global perspectives. To learn content and style conditions simultaneously, we introduce a StyDiT Block for transformer-based diffusion to learn content-style disentanglement from an image. Current artistic image datasets, like Wikiart~\cite{wikiart}, often do not provide clear Canny information. This limitation hinders their applicability in diffusion model-based methods that necessitate the simultaneous training of Canny as a content condition alongside style conditions. To overcome this challenge, we introduce a new ultra-high style image dataset, Aes4M, which encompasses 10 categories and 4 million highly aesthetic and artistic images. This dataset is characterized by its high aesthetic quality, consistent text-image alignment, and clear Canny conditions.

\textbf{Limitation.}
Although our proposed U-StyDiT can generate ultra-high quality artistic stylized images, we are limited by manpower and computational resources. This prevents us from obtaining a dataset with hundreds of thousands of style categories, each containing over 500 images with clear Canny information to validate our method. This limitation means our approach cannot support arbitrary style transfer in an open-world setting.

{
    \small
    \bibliographystyle{ieeenat_fullname}
    \bibliography{main}
}

\clearpage
\maketitle
\setcounter{section}{0}
\section{The pipline of inference.}
\begin{figure}[htb]
		\centering
		\includegraphics[width=1\columnwidth]{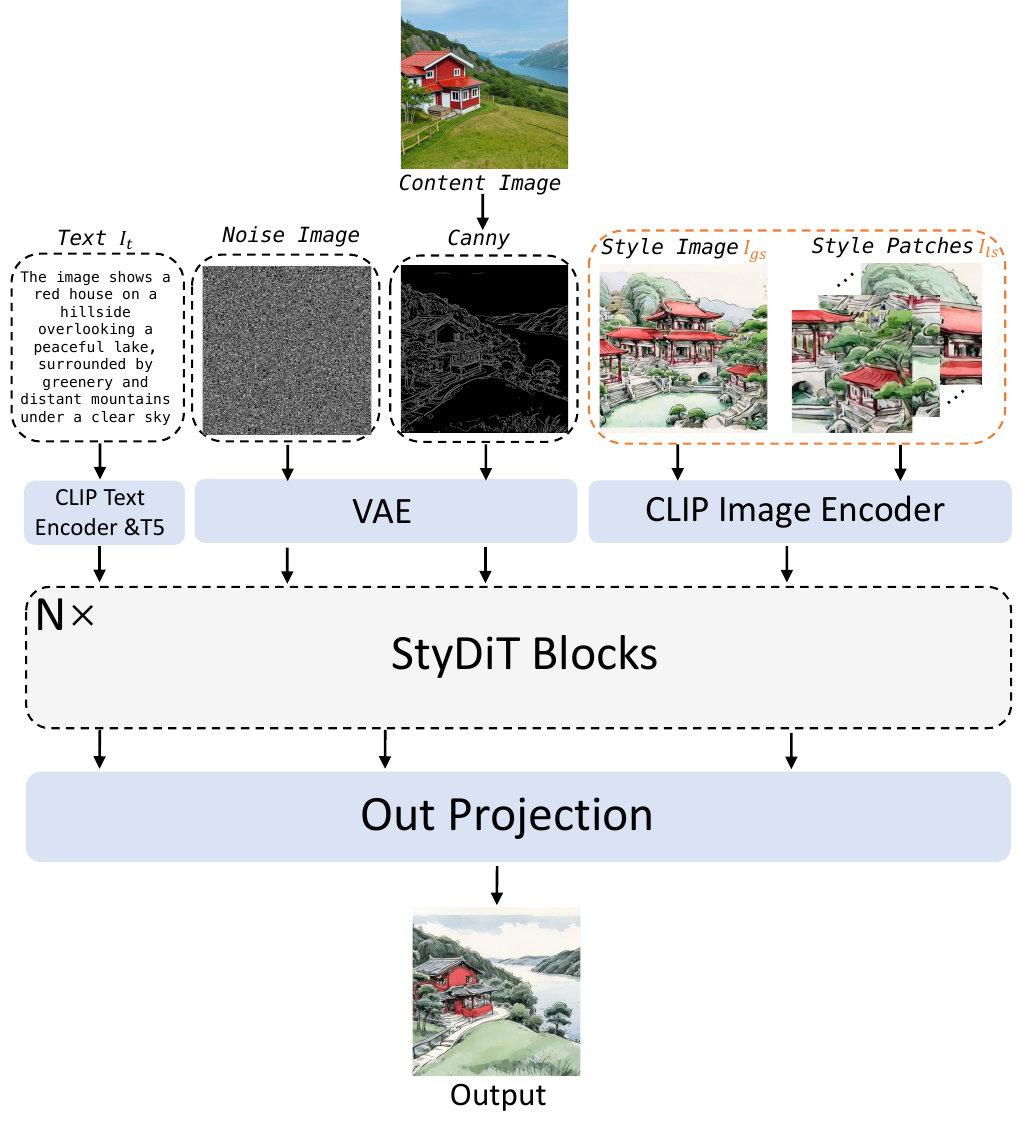} 
		\caption{The pipline of inference.
		}
		\label{image13}
\end{figure}

As shown in Fig.~\ref{image13}, we present the inference process of the proposed U-StyDiT. We collect some high-quality image from~\cite{flickr} as content image. Given a content image, we first extract the Canny image~\cite{canny1986computational} as content condition and use Chatglm~\cite{glm2024chatglm} to extract a text description. 
If the given style image has a high resolution, such as $1024\times 1024$, we resize it to $512\times 512$ and extract multiple $512\times 512$ style image patches as style conditions. If the given style image has a lower resolution, below $512\times 512$, we resize it to $512\times 512$ and obtain multiple style image patches through duplication to serve as style conditions. Finally, this information is then fed into U-StyDiT to generate the desired stylized image.

\section{The additional details of Aes4M.}
\begin{figure*}[htb]
		\centering
		\includegraphics[width=2.05\columnwidth]{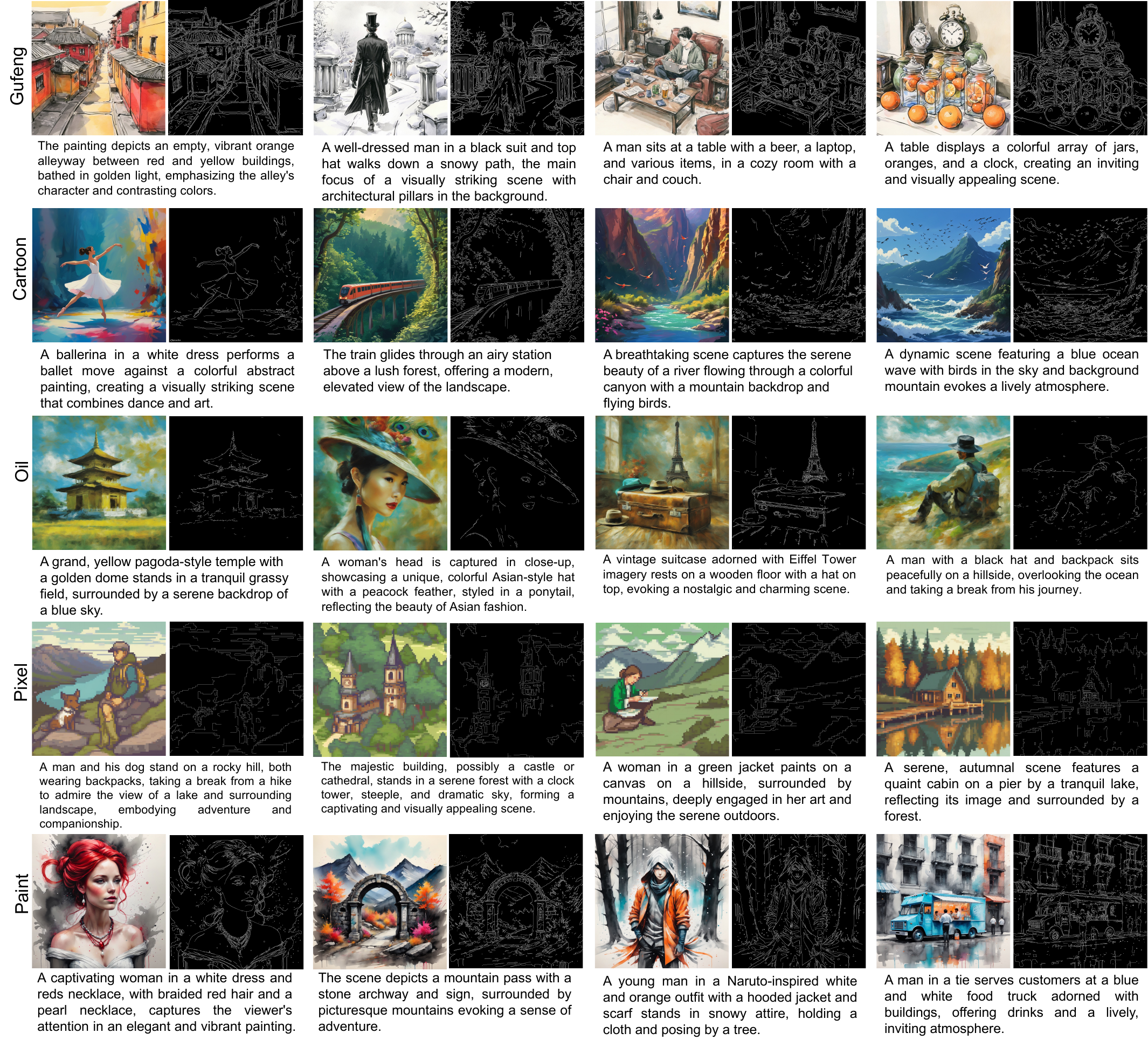} 
		\caption{We present some style image examples with Gufeng, cartoon, oil, pixel, and paint styles.
		}
		\label{image14}
\end{figure*}

\begin{figure*}[htb]
		\centering
		\includegraphics[width=2.05\columnwidth]{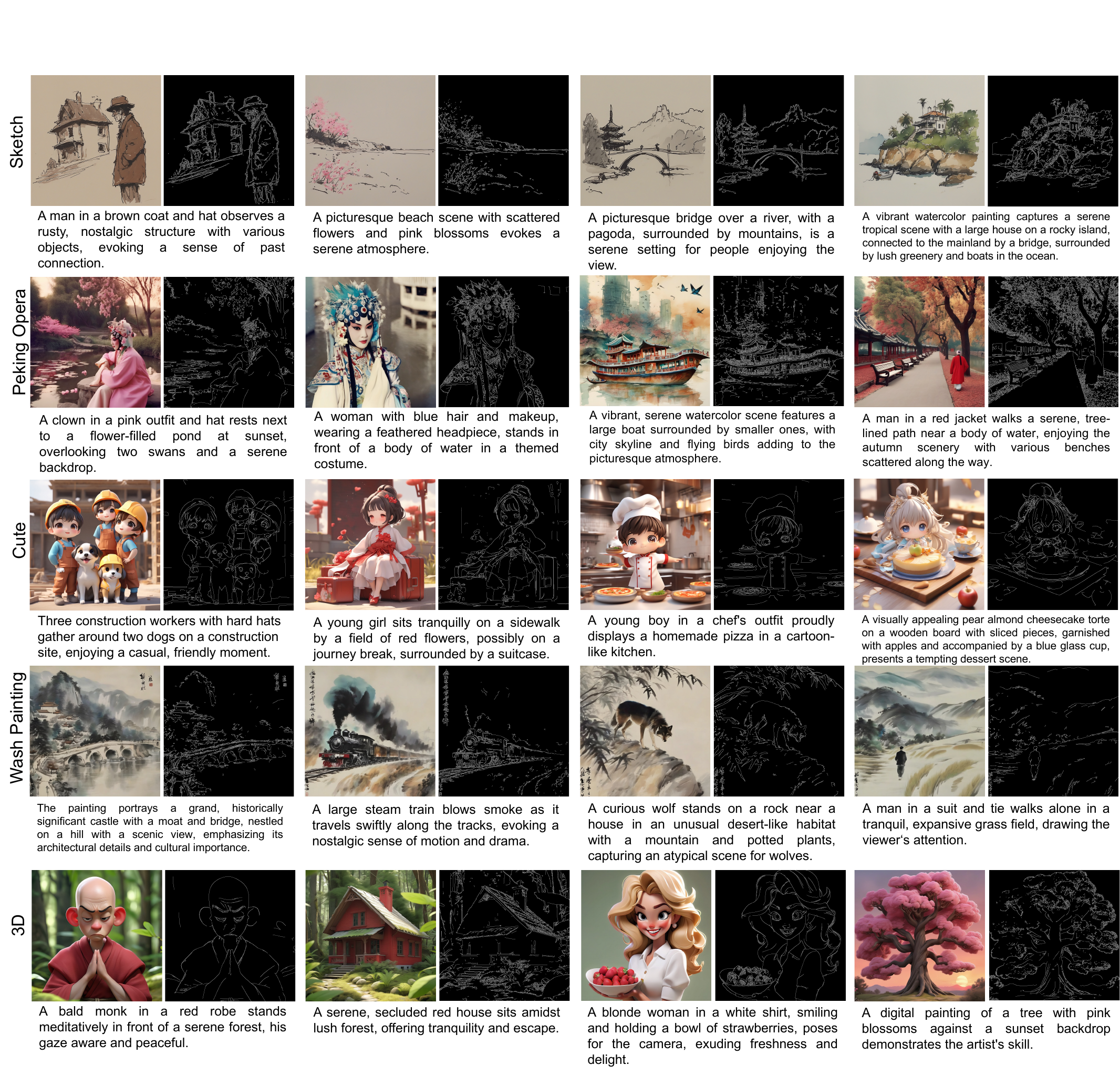} 
		\caption{We show some style image examples with sketch, Peking opera, cute, wash painting, and 3D styles. 
		}
		\label{image15}
\end{figure*}

Aes4M consists of 10 categories, with 400,000 artistic images in each category, totaling 4 million. Each image has a resolution of $1024\times1024$. As shown in Fig.~\ref{image14} and Fig.~\ref{image15}, we show some style image examples, including canny images (i.e., the high threshold is set to 200, and the low threshold is set to 100) and text descriptions.

\end{document}